\documentclass[10pt]{article}
\usepackage[preprint]{tmlr}
\usepackage{amsmath,amssymb,amsfonts}
\usepackage{graphicx}
\usepackage{booktabs}
\usepackage[hidelinks]{hyperref}
\usepackage{url}
\usepackage{microtype}
\hypersetup{
  pdftitle={What You Can't See Is Still What You Learn: A Preregistered Sixty-Society Confirmation That Evidence Masking Drives Compositional Generalization},
  pdfauthor={Narcis Marincat},
  pdfsubject={Preregistered confirmation of the evidence-masking effect on compositional generalization in shared-genome language-model societies}
}
\newcommand{\soc}[1]{\texttt{#1}}
\title{What You Can't See Is Still What You Learn:\\ A Preregistered Sixty-Society Confirmation That Evidence Masking Drives Compositional Generalization}
\author{Narcis Marincat \\ Independent researcher \\ Correspondence: \texttt{narcis.marincat.20@ucl.ac.uk}}

\begin{document}
\maketitle

\begin{abstract}
Restricting what a module can read may improve what a system learns to compute. We test this in a preregistered confirmation with sixty four-cell systems sharing a frozen language-model backbone and communicating through learned continuous packets. Five conditions vary evidence masking, ownership markers, and replacement of foreign evidence with neutral filler (task-irrelevant text of the same token length), across six initialization clusters, each with two data orders, on one fresh task world. With markers available in both regimes, masking improved accuracy on held-out two- and three-operation compositions by median paired differences of 0.846 and 0.859; all twelve pairs cleared the required margins, and the full preregistered behavioral criterion passed. The unmarked replication also passed. No marked global-visibility (G$+$) system passed the marker-following check, so the effect of usable role information remains unresolved. The filler condition yielded seven full generalizers, but its decomposition criteria were inconclusive. Packet interventions in all eighteen audited masked systems followed the predicted intermediate-value changes on eligible cases; these finite, success-conditioned audits do not establish mediation. The results confirm a large advantage of the tested masking regime, while leaving its finer attribution and generality open. Protocols, results, and checkpoints are public.
\end{abstract}

\section{Introduction}\label{sec:intro}

Giving every module the full input seems helpful: each has all the evidence needed to solve the task. But it also permits solutions that bypass the intended division of work. Restricting direct evidence access may instead favor a system that passes reusable intermediate results between modules. The empirical question is whether such a restriction improves generalization when familiar operations must be combined in previously unseen programs.\footnote{The author conceived the research ideas and experimental design, developed the overall workflow, and personally verified the reported results. Generative AI assisted implementation, pipeline operation and monitoring, result reporting and verification checks, protocol refinement and review, and manuscript drafting and revision. Further details appear in the generative-AI disclosure before the references.}

We study a deliberately small setting where both performance and the communicated state can be tested precisely. Four sequential calls to one shared language model form a \emph{society}: the first cell receives a starting value, and the remaining cells are assigned operations or forwarding instructions. They communicate through learned continuous vectors. Under masking, a cell can read only its assigned evidence span, the shared question, and its incoming packet; under global visibility, it can also read the other spans. These are jointly trained modules, not independently acting agents. The task is to compute ordered compositions of familiar arithmetic operations expressed in templated language.

The parent study \citep{marincat2026populus} found large masking advantages, but its preregistered battery formally failed: median depth-three accuracy was 0.6988 against a 0.70 floor. It also left an attribution question open. Masking both removes foreign evidence and identifies which evidence belongs to a cell. The present five-condition study tests the masking advantage on a fresh task world while varying ownership markers (labels identifying a cell's own evidence) and neutral filler (task-irrelevant text replacing other cells' evidence). A separate marker-following check determines whether marker availability actually supplies usable role information.

\paragraph{Contributions.}
\begin{enumerate}
\item \textbf{A large masking advantage under the frozen confirmation criteria.} All twelve matched pairs exceeded the required margin at both depths with markers available (median paired advantages $+0.846$/$+0.859$); the unmarked replication also passed. The sixty trajectories comprise six initialization clusters, each with two orders per condition, not sixty independent task worlds.
\item \textbf{Controls that bound the explanation.} Markers met the practical-redundancy criterion under masking, but no marked global-visibility (G$+$) trajectory passed the marker-following check. Semantic role rescue therefore remains unresolved. Replacing other cells' evidence with neutral filler of the same token length yielded seven full generalizers, yet the preregistered contrasts intended to distinguish the competing explanations were inconclusive.
\item \textbf{Value-level intervention evidence with explicit limits.} All eighteen audited masked trajectories passed the same-value and counterfactual packet tests perfectly on their eligible cases. A same-value test substitutes a donor packet associated with the same intermediate arithmetic value and checks that the answer is preserved. A counterfactual test substitutes a donor packet associated with a different intermediate value and checks that the remaining computation produces the answer predicted from that replacement. Marked global trajectories did not meet the strong-canonicality criterion across all interfaces. These audits characterize intervention behavior; they do not establish that a particular representation mediates generalization.
\end{enumerate}

The protocol and numerical decision rules were frozen before seed generation. Execution departed from the planned cohort-wide evaluation embargo, and some admission and duplicate-audit records were not retained. These limitations are disclosed in \S\ref{sec:integrity} and Appendices~\ref{app:integrity}--\ref{app:correction}; passing the numerical criteria is not a claim of flawless protocol execution. Throughout, ``masking advantage'' refers to the tested architecture, task world, and training budget.

\section{Related work}\label{sec:related}

Compositional generalization is the ability to use familiar components in combinations not seen during training. \citet{lake2018generalization} introduced SCAN, a benchmark that maps short commands such as ``jump twice'' to sequences of actions. Their models could succeed when training and test commands were similar, yet struggle when familiar elements had to be combined in new ways. Here, the corresponding question is whether a system can execute previously unseen sequences of familiar arithmetic operations.

A separate question concerns the structure of the messages used to solve a task. In \emph{emergent communication}, agents learn a communication code while learning to cooperate. A code is \emph{compositional} when its parts have reusable meanings and combine systematically. \citet{kottur2017natural} showed that successful cooperation need not produce such a code, and that restrictions on communication can encourage it. \citet{chaabouni2020compositionality} likewise found that agents could handle unseen combinations without strongly compositional messages. More compositional codes were, however, easier for new receivers to learn: a trained sender was kept fixed while a new receiver learned to interpret its messages. This separates generalization by an existing system from the ease of teaching its code to another learner.

Our packet tests address another distinction: correct final answers do not by themselves reveal what a message does inside the computation. \citet{geiger2021causal} use \emph{interchange interventions} to test such hypotheses. An internal state produced for one input replaces the corresponding state for another input; the resulting output is compared with the prediction from an equivalent substitution in a simpler, symbolic computation. We similarly replace packets and test whether the output follows the predicted change in an intermediate arithmetic value. Our tests cover finite sets of eligible, initially correct cases. They do not establish the same correspondence for every possible input, or show that it explains the generalization advantage from masking.

Building on the parent study \citep{marincat2026populus}, we test evidence masking on a newly generated arithmetic task world in a preregistered, four-cell system with shared model parameters and continuous-vector messages. Ownership labels and replacement of other cells' evidence with neutral text help assess competing explanations. The companion study instead examines whether learned communication interfaces transfer between systems and tasks \citep{marincat2026companion}.

\section{Experimental design}\label{sec:design}

\paragraph{Task and a worked example.} A task world contains twelve distinct non-identity affine bijections $f_i(x)=(a_i x+b_i)\bmod17$, with $a_i\ne0$. The target for a depth-$d$ program is $f_{i_d}(\cdots f_{i_1}(x)\cdots)$, reported as one of seventeen letter labels. For example, one held-out depth-three program in this world applies $(5x+6)$, $(5x+6)$, and $(3x+12)$, all modulo 17, to the starting value 2:
\[
\underbrace{2}_{\text{cell 1: start}}
\xrightarrow{\;5x+6\;}
\underbrace{16}_{\text{cell 2}}
\xrightarrow{\;5x+6\;}
\underbrace{1}_{\text{cell 3}}
\xrightarrow{\;3x+12\;}
\underbrace{15}_{\text{cell 4: final value}}.
\]
The arrows show the intended mathematical computation, not supervised packet targets. Cells receive natural-language descriptions, such as the sealed phrasing ``The operation specified for this slot is to compute 5 times the input plus 6, wrapping around at 17.'' embedded in the frozen grammar. The packet protocol and local computation are learned from the final-answer objective. At depth two, the last cell receives a forwarding instruction; identity and single-operation episodes use forwarders in the unoccupied operation slots.

\paragraph{Architecture and communication.}
Each society uses four sequential calls to a frozen Qwen2.5-0.5B-Instruct backbone \citep{qwen25}, with one shared rank-8 LoRA adapter \citep{hu2022lora} and no cell-specific learned parameters. The reader, writer, and readout are also shared within a society; different societies have separately trained parameters. Each packet contains two 896-dimensional vectors. A shared linear reader maps the incoming packet into two pseudo-token embeddings; a shared linear writer maps the corresponding output states into packet updates $\Delta P$. With normalization $N(v)=v/\sqrt{\max(\operatorname{mean}(v^2),0.0025)}$ applied separately to each vector, the update is $P=N(U+0.05\,\Delta P)$, where $U$ is the normalized incoming packet. This residual anchor carries the incoming message directly into the update.

At the source, $U=0$ and incoming packet positions are unavailable for attention to read, although their output states are used to write the source packet. Downstream packet positions are normally readable. The all-cut intervention disables their readability and removes the residual anchor; zero-packet replacement retains readability while substituting zero-valued vectors. These interventions are distinct.

\paragraph{Output and trainable parameters.}
A learned readout projection (the \emph{mouth}) maps the flattened fourth-cell packet into the frozen language-model head's input space:
\[
h_{\mathrm{final}}=h_{\mathrm{base}}+0.003\,W_{\mathrm{mouth}}\operatorname{vec}(P_4),
\qquad
\widetilde z_j=z_j(h_{\mathrm{final}})-z_j(h_{\mathrm{base}}).
\]
Here $h_{\mathrm{base}}$ comes from a separate question-only backbone call with LoRA disabled, and $z_j$ is the frozen head's logit for label $j$. Both training and evaluation use these residualized logits over seventeen fixed answer tokens; evaluation selects their argmax. The head is linear, so the baseline cancels algebraically, leaving a classifier on the final packet rather than unrestricted text generation. All communication/readout projections are bias-free. A bias-free auxiliary classifier predicts the final label from the last packet during training only; it does not impose intermediate-value targets. Including it, the system has 4{,}323{,}072 trainable parameters (1{,}081{,}344 LoRA and 3{,}241{,}728 projection/auxiliary parameters).

\paragraph{Training and held-out evaluation.} Training samples identity, single-operation, depth-two, and depth-three episodes with probabilities $0.10$, $0.25$, $0.325$, and $0.325$. All twelve primitives are trained; the composition banks contain 91 depth-two and 401 depth-three programs. A singleton operation is placed uniformly in one of the three operation slots. Starting values and eligible programs are sampled uniformly, with 24 training phrasings per operator. The fresh world is generated after the freeze, excluding development-exposed seeds. Held-out programs are disjoint from training and monitoring programs, and their complete affine maps occur in neither the composition curriculum nor the identity/single-operation curriculum. The primary banks contain 18 depth-two and 231 depth-three programs, each evaluated at all 17 values and four sealed phrasing rotations: 1{,}224 and 15{,}708 cases per trajectory. These are fixed-bank measurements, not independent experimental replicates. The single-operation diagnostic contains $12\times3\times17\times4=2{,}448$ cases. Held-out phrasings are within the same generated grammar, not free-form language.

Each run completes 20{,}000 AdamW updates at batch size 32 (two accumulated microbatches of 16), in float32 with gradient clipping at norm 1.0.\footnote{The frozen schedules are evaluated using the zero-based pre-update counter $t=0,\ldots,19{,}999$; the first optimizer call therefore has zero scheduled learning rate, and the final call approaches the nominal 10\% endpoint.} Projection/auxiliary parameters use learning rate $5\times10^{-4}$ and weight decay 0.01; LoRA uses $10^{-4}$ and zero weight decay. Rates warm up linearly over 1{,}000 updates, remain constant through update 5{,}000, and decay by a cosine schedule to 10\% at update 20{,}000. LoRA targets the attention query, key, value, and output projections, with rank 8, alpha 16, and zero dropout. The loss is final-label cross-entropy plus auxiliary cross-entropy with weight 0.5 through update 4{,}000, linearly annealed to zero by update 10{,}000. Only the final checkpoint is evaluated for the reported outcomes.

\paragraph{Five conditions.} The design crosses evidence-access regime with explicit span-ownership marking:
\begin{itemize}
\item \textbf{R$-$} --- \emph{masked, unmarked}: each cell attends only to its own evidence span (the parent's restricted regime under the revised grammar); each span carries the neutral two-token header \texttt{slot:} (the same two-token block that holds \texttt{mine:}/\texttt{other:} in the marked arms, keeping token layout identical).
\item \textbf{R$+$} --- \emph{masked, marked}: as R$-$, but every span carries an explicit ownership marker (`` mine:'' on the cell's own span, `` other:'' on foreign spans).
\item \textbf{G$-$} --- \emph{global, unmarked}: all spans directly readable, each carrying the neutral two-token header \texttt{slot:} (the parent's global regime).
\item \textbf{G$+$} --- \emph{global, marked}: all spans readable, all spans ownership-marked.
\item \textbf{N$+$} --- \emph{neutral-filler, marked}: global attention geometry, but every foreign evidence span is replaced by token-length-matched, task-irrelevant neutral filler drawn by a committed pseudorandom function from a frozen bank; markers present. N$+$ gives a cell the global regime's attention reach without the original foreign evidence tokens, while preserving span lengths; because filler replacement also changes lexical content and may make the sole genuine evidence span recognizable as such, it informs, but does not isolate, the contribution of foreign content (\S\ref{sec:limits}).
\end{itemize}
Marker headers are constant strings within spans; under masking they are positionally redundant during training (a cell's own span is always the same slot), a property we return to in \S\ref{sec:discussion}.

\paragraph{Matched-twin discipline.} The committed tier (GOLD) is six initialization clusters $\times$ two orders per initialization $\times$ five conditions $=60$ societies. Within a cluster/order, all five conditions share initialization bytes, the semantic training-example stream (hash-verified; serialized inputs differ across conditions exactly as the conditions prescribe), token layout, positional geometry, trainable parameterization, and the 20{,}000-update budget; only the mask, markers, or filler differ. Final checkpoints only; no checkpoint selection.

\paragraph{Preregistered contrasts.} Signed paired contrasts per matched trajectory: C1 $=$ R$+ -$ G$+$ (primary behavioral), C2 $=$ G$+ -$ G$-$ (marker rescue of the global regime), C3 $=$ R$+ -$ R$-$ (marker redundancy under masking), C4 $=$ R$+ -$ N$+$ (neutral-context exposure), C5 $=$ N$+ -$ G$+$ (task-relevant foreign content), C6 $=$ R$- -$ G$-$ (parent replication).

A complete directional pass requires the full frozen conjunction: $\geq n{-}1$ of $n{=}12$ pairs with signed margins $\geq 0.20$ at both depths; median $\geq 0.25$ at both depths; advantaged-arm median depth-three accuracy $\geq 0.70$ with no run below $0.40$ at either depth; and all-packets-cut accuracy $\leq 0.11$ in $\geq n{-}1$ advantaged-arm trajectories. C3 has a practical-redundancy criterion instead of a direction.

A mandatory manipulation check (\textsc{gplus\_tag\_use}) requires 11/12 G$+$ societies to reach $\geq0.90$ on \emph{both} intact-marker accuracy and causal counterfactual marker tracking under sealed marker permutations before any ``role-equalized'' reading of C1 is permitted; the frozen reporting separates \textsc{c1\_behavioral}, \textsc{gplus\_tag\_use}, and \textsc{primary\_role\_equalized}, and no omnibus cohort pass exists.

Confirmatory decisions use the frozen conjunctions; descriptive uncertainty summaries retain initialization clustering (the exact sign-flip test flips the sign of whole initialization clusters' mean differences to build its null distribution; with six clusters, the smallest attainable two-sided $p$ is $1/32=0.03125$). These tests assume sign symmetry of the cluster effects under the null; they are descriptive paired tests, not design-based randomization tests.

\section{Preregistration and execution}\label{sec:integrity}

The protocol, code inventory, contrast definitions, and decision rules were deposited before seed generation, with third-party Software Heritage snapshots; the populated manifest was deposited before training. The intended machine-admission checks, five AI-assisted protocol-review rulings, running stream hashes, and exclusive-lock verdict process are described in Appendix~\ref{app:integrity}. The AI reviews were procedural assistance, not journal peer review or independent replication.

Two distinctions matter for interpreting the confirmation. First, a post-evaluation, pre-verdict correction changed only the manifest's derived single-operation denominator from 1{,}224 to 2{,}448; the frozen evaluator and sixty result records were unchanged. Second, the pipeline evaluated each final checkpoint as it completed, violating the planned embargo until all sixty had finished. Consequently, outcomes were inspected before the correction and verdict. No outcome-dependent scientific protocol change is reported, but the intended protection against early-outcome influence was not maintained. Missing host-admission transcripts and unretained duplicate-audit outputs further limit retrospective verification. Appendix~\ref{app:correction} preserves the full correction and timing disclosures.

\begin{figure}[t]
\centering
\includegraphics[width=0.95\textwidth]{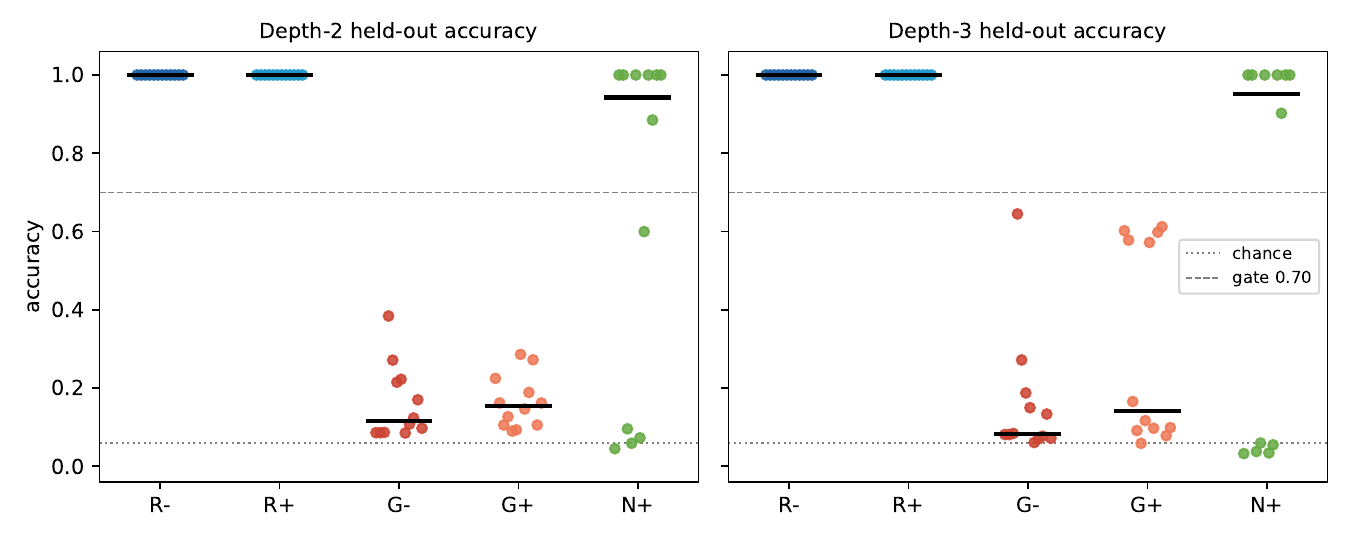}
\caption{\textbf{Held-out compositional accuracy by arm.} Depth-two and depth-three accuracy for all sixty societies (six initialization clusters $\times$ two orders per arm). Masked arms (R$-$, R$+$) cluster at ceiling; global arms (G$-$, G$+$) are low but heterogeneous (five G$+$ and one G$-$ trajectories are deep-only under the frozen classifier); the neutral-filler arm (N$+$) splits into seven full generalizers, four non-acquirers, and one shallow-partial trajectory. Masked trajectories are exactly at chance ($0.0588$) under full packet severance; other competent trajectories are near chance and within the frozen bound.}
\label{fig:arms}
\end{figure}

\begin{figure}[t]
\centering
\includegraphics[width=0.95\textwidth]{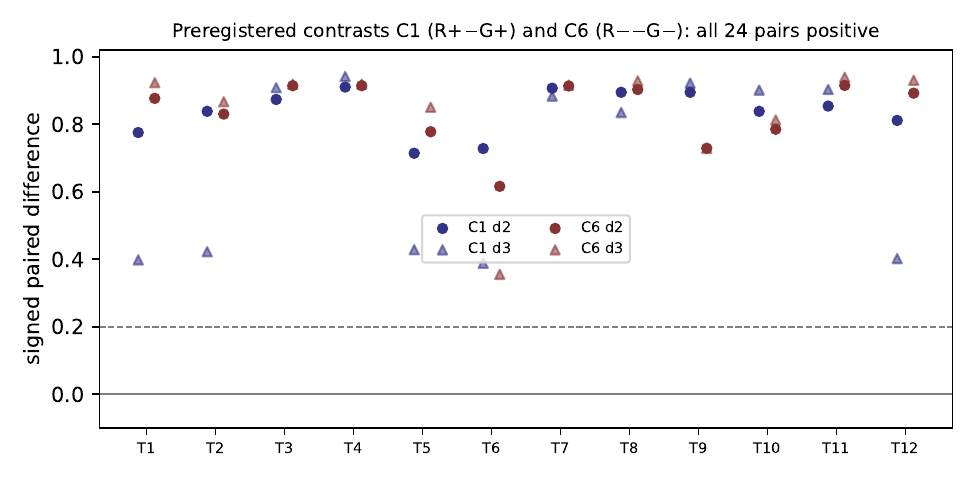}
\caption{\textbf{Primary behavioral contrast and preregistered replication.} Signed per-pair differences at depths two and three for C1 (R$+{-}$G$+$) and C6 (R$-{-}$G$-$) across the twelve matched trajectories (T1--T12 index the twelve initialization--order sets), against the frozen 0.20 per-pair margin (dashed). All 24 paired comparisons at both depths clear the margin.}
\label{fig:contrasts}
\end{figure}

\section{Preregistered results}\label{sec:results}

Table~\ref{tab:arms} summarizes the arms; Table~\ref{tab:verdict} the gate outcomes; Table~\ref{tab:census} the frozen phenotype census; Table~\ref{tab:secondary} the preregistered secondary statistics. Provenance: behavioral records are from the locked compiled bundle; gate decisions from the one-shot verdict archive; phenotypes and secondary statistics from the frozen classifier and statistics engine applied to the locked bundle; mechanism results from the later non-gating audit deposit.

\begin{table}[t]
\centering
\setlength{\belowcaptionskip}{8pt}
\caption{\textbf{Arm-level medians} over twelve societies per condition. $A_2$ and $A_3$ are accuracies on held-out compositions of two and three operations, respectively. The $A_2\geq0.70$ column counts societies reaching 70\% accuracy at depth two; the full-generalizer category additionally requires $A_3\geq0.70$ (Table~\ref{tab:census}). Median all-cut reports how much depth-three accuracy remains when all inter-cell communication is severed during evaluation of the trained checkpoint, with $1/17$ as the reference chance level.}
\label{tab:arms}
\begin{tabular}{lcccc}
\toprule
Arm & Median $A_2$ & Median $A_3$ & $A_2\geq0.70$ & Median all-cut \\
\midrule
R$-$ (masked) & 1.000 & 1.000 & 12/12 & 0.059 \\
R$+$ (masked, marked) & 1.000 & 1.000 & 12/12 & 0.059 \\
G$-$ (global) & 0.116 & 0.083 & 0/12 & 0.060 \\
G$+$ (global, marked) & 0.154 & 0.141 & 0/12 & 0.057 \\
N$+$ (neutral-filler, marked) & 0.942 & 0.951 & 7/12 & 0.059 \\
\bottomrule
\end{tabular}
\end{table}

\begin{table}[t]
\centering
\setlength{\belowcaptionskip}{8pt}
\caption{\textbf{Preregistered gate outcomes} from the one-shot locked verdict. PASS means all requirements for the stated criterion were met. FAIL indicates an unmet criterion and does not establish the absence of an effect. INCONCLUSIVE means the preregistered decision rules did not resolve the intended comparison. For C4/C5, this label summarizes separate archived outcomes: both directional criteria failed, and both practical-null evaluations were uninformative because their competence requirements were not met. For C1, C2, and C6, pair counts report how many matched pairs met the required accuracy margins at both depths. Paired medians are accuracy differences reported as depth-two/depth-three. C3 is labelled ``criterion met'' here; the machine verdict calls it \texttt{PRACTICAL\_NULL}, meaning practical redundancy at the frozen tolerance, not statistical equivalence. These are numerical gate outcomes; execution deviations and provenance limitations are reported separately (\S\ref{sec:integrity}, \S\ref{sec:limits}, Appendix~\ref{app:correction}).}
\label{tab:verdict}
\footnotesize
\begin{tabular}{llll}
\toprule
Gate & Preregistered claim & Result & Key numbers \\
\midrule
C1 & Masking advantage (marked) & \textbf{PASS} & 12/12 pairs, median $+0.85/+0.86$ \\
C6 & Parent replication (unmarked) & \textbf{PASS} & 12/12 pairs, median $+0.88/+0.92$ \\
C3 & Marker redundancy (masked) & \textbf{CRITERION MET} & 12/12 in bound, median $0.000$ \\
Tag use & G$+$ marker following & \textbf{FAIL} & 0/12 pass; tracking 0.038--0.060 \\
C2 & Marker rescue (global) & \textbf{FAIL} & 0/12 pairs, median $+0.01/+0.03$ \\
C4/C5 & N$+$ decomposition & \textbf{INCONCLUSIVE} & N$+$: 7 full gen.\ / 4 non-acq.\ / 1 shallow \\
\bottomrule
\end{tabular}

\end{table}

\begin{table}[t]
\centering
\setlength{\belowcaptionskip}{8pt}
\caption{\textbf{Frozen phenotype census} (classifier and priority rules frozen with the preregistration; applied to the locked bundle). Counts of the twelve societies in each condition, grouped by their training and held-out performance. Categories, with $A_d$ the held-out compositional accuracy at depth $d$ and $T$ the better of the two training-bank fit scores: \emph{full generalizer} --- $A_2\geq0.70$ and $A_3\geq0.70$; \emph{deep-only} (abbreviating \emph{deep-only generalization inversion}) --- $A_3\geq0.40$ with $A_2<0.40$ and a gap of at least $0.20$, i.e.\ higher held-out accuracy at depth three than at depth two; \emph{shallow-partial} --- $A_2\geq0.40$ with $A_3<0.40$; \emph{memorizer} --- fits the training bank ($T\geq0.90$) yet both held-out accuracies below $0.20$; \emph{non-acquirer} --- fails even the training fit ($T<0.90$) with both held-out accuracies below $0.20$; \emph{mixed} (abbreviating \emph{intermediate-mixed}) --- any remaining profile.}
\label{tab:census}
\begin{tabular}{lcccccc}
\toprule
Arm & Full gen. & Deep-only & Shallow-partial & Memorizer & Non-acquirer & Mixed \\
\midrule
R$-$ & 12 & 0 & 0 & 0 & 0 & 0 \\
R$+$ & 12 & 0 & 0 & 0 & 0 & 0 \\
G$-$ & 0 & 1 & 0 & 8 & 0 & 3 \\
G$+$ & 0 & 5 & 0 & 7 & 0 & 0 \\
N$+$ & 7 & 0 & 1 & 0 & 4 & 0 \\
\bottomrule
\end{tabular}

\end{table}

\begin{table}[t]
\centering
\setlength{\belowcaptionskip}{8pt}
\caption{\textbf{Preregistered secondary statistics.} Means and medians summarize the twelve paired accuracy differences for each comparison and depth. The 95\% confidence intervals concern the mean difference. We calculate these intervals by repeatedly sampling six initialization groups with replacement, keeping the two training streams within each group together---a cluster bootstrap with 100{,}000 resamples and random seed 271828. The $p$-values use the six-group sign-flip test described in Section~\ref{sec:design}. Intervals and $p$-values are reported separately for each comparison, without correction for multiple comparisons. These analyses supplement the preregistered decision criteria in Table~\ref{tab:verdict}; they do not determine whether those criteria pass. The preregistered interaction $I_d=\mathrm{C2}_d-\mathrm{C3}_d$ is reported in the last rows ($I_3$ is not numerically identical to C2 because one R$+$/R$-$ trajectory has a small nonzero depth-three difference). Trajectory-keyed differences under a single labelling convention, initialization means, the full interaction and decomposition outputs, and the frozen initialization-grouped phenotype report are in the released \texttt{SECONDARY\_STATS\_v2.json}. Note the trajectory-level identities $\mathrm{C1}_d=\mathrm{C4}_d+\mathrm{C5}_d$ and $I_d=\mathrm{C2}_d-\mathrm{C3}_d$ hold per trajectory (maximum absolute trajectory-level decomposition residual $1.11\times10^{-16}$); medians do not obey these identities, so no component share is inferred from medians.}
\label{tab:secondary}
\footnotesize
\begin{tabular}{lccccc}
\toprule
Contrast & Depth & Median & Mean & 95\% CI for mean & Sign-flip $p$ \\
\midrule
C1 (R$+{-}$G$+$) & 2 & $+0.846$ & $+0.837$ & $[+0.784,\ +0.881]$ & 0.03125 \\
                 & 3 & $+0.859$ & $+0.694$ & $[+0.524,\ +0.864]$ & 0.03125 \\
C2 (G$+{-}$G$-$) & 2 & $+0.007$ & $+0.002$ & $[-0.047,\ +0.044]$ & 0.9375 \\
                 & 3 & $+0.034$ & $+0.146$ & $[-0.014,\ +0.313]$ & 0.1875 \\
C3 (R$+{-}$R$-$) & 2 & $0.000$ & $0.000$ & $[0.000,\ 0.000]$ & 1.0 \\
                 & 3 & $0.000$ & $-3.71 \times 10^{-5}$ & $[-1.11 \times 10^{-4},\ 0]$ & 1.0 \\
C4 (R$+{-}$N$+$) & 2 & $+0.058$ & $+0.354$ & $[+0.119,\ +0.626]$ & 0.0625 \\
                 & 3 & $+0.049$ & $+0.407$ & $[+0.169,\ +0.644]$ & 0.03125 \\
C5 (N$+{-}$G$+$) & 2 & $+0.736$ & $+0.483$ & $[+0.177,\ +0.739]$ & 0.0625 \\
                 & 3 & $+0.351$ & $+0.288$ & $[-0.081,\ +0.594]$ & 0.21875 \\
C6 (R$-{-}$G$-$) & 2 & $+0.884$ & $+0.839$ & $[+0.768,\ +0.900]$ & 0.03125 \\
                 & 3 & $+0.917$ & $+0.841$ & $[+0.735,\ +0.921]$ & 0.03125 \\
\midrule
$I$ (C2$-$C3)    & 2 & $+0.007$ & $+0.002$ & $[-0.047,\ +0.044]$ & 0.9375 \\
                 & 3 & $+0.034$ & $+0.146$ & $[-0.014,\ +0.313]$ & 0.1875 \\
\bottomrule
\end{tabular}

\end{table}

\paragraph{The masking contrasts pass their complete conjunctions (C1, C6).} With markers present in both arms, all twelve matched R$+$/G$+$ trajectories exceed the 0.20 margin at both depths (smallest observed margins: $+0.71$ at depth two, $+0.39$ at depth three, in different trajectories); the median paired advantage is $+0.846$ at depth two and $+0.859$ at depth three; the advantaged arm's median depth-three accuracy is 1.000 with no run below 0.40; and 12/12 advantaged trajectories are within the all-cut bound (each exactly at chance, 0.0588). The unmarked replication C6 likewise passes in full: 12/12 trajectories, median $+0.884$/$+0.917$. The parent effect, a formal miss by 0.0012 under its original battery, thus passes the separately preregistered confirmation criteria on a fresh sealed world (the confirmation required at least eleven of twelve trajectories to clear both directional margins; all twelve did so for C1 and C6); the parent result itself remains a formal miss under its original battery. C1's pre-committed behavioral verdict (\textsc{c1\_behavioral}) is PASS. These are twenty-four contrast-specific paired comparisons across twelve matched initialization--order sets; the marked and unmarked comparisons reuse the same initialization--order structure and are not statistically independent replicates.

\paragraph{Markers meet the practical-redundancy criterion under masking (C3).} The signed R$+{-}$R$-$ difference has median exactly $0.000$ at both depths; all twelve trajectories are within the $\pm0.20$ bound, the median absolute difference is $0.000$ against the $0.10$ tolerance, and both arms meet the competence floor. R$+$ and R$-$ therefore met the preregistered practical-redundancy criterion at the frozen resolution. This is satisfaction of the preregistered practical-redundancy criterion, not a formal statistical equivalence test or a causal assay of internal marker use.

\paragraph{No marked global-visibility (G$+$) trajectory met the marker-following criterion (\textsc{tag use}); no gate-level rescue (C2).} The mandatory G$+$ manipulation check requires, per model, intact-marker accuracy $\geq0.90$ and counterfactual tracking $\geq0.90$ under sealed marker permutations (model-level quantities are unweighted means over the five permutation-specific fidelities). No G$+$ trajectory met it: counterfactual tracking ranges 0.038--0.060 and intact-marker accuracy 0.053--0.525. Consistently, C2 fails its complete conjunction: zero of twelve trajectories clear the required margin at both depths (median $+0.007$/$+0.034$). Four trajectories show depth-three improvements exceeding 0.40 ($+0.44$, $+0.52$, $+0.42$, $+0.53$), distributed across three initialization clusters; none clears the required 0.20 margin at both depths. Per the frozen interpretation rule, the conjunction ``masking gap persists after functional role equalization'' (\textsc{primary\_role\_equalized}) is \emph{not} claimed: no G$+$ trajectory demonstrated reliable causal marker following, so C1 stands as a marker-present behavioral comparison, and semantic role rescue remains unresolved rather than refuted. A follow-up design that makes markers informative --- shuffling slot assignments so position no longer predicts ownership --- is documented (\S\ref{sec:followups}); functional role following would still require validation there.

\paragraph{The neutral-filler arm splits three ways (C4, C5).} Under the frozen classifier, seven N$+$ trajectories are full generalizers (arm medians $A_2$/$A_3$ $=0.942$/$0.951$; one full generalizer scores $0.885$/$0.902$, so the category is not a synonym for ceiling), four are non-acquirers, and one is shallow-partial (\soc{m999254511\_o2610306098}: $A_2=0.600$, $A_3=0.055$) --- N$+$ separates strongly at depth three, with a shallow-partial exception to any simple success-versus-collapse description across both depths. The frozen competence preconditions for the C4 and C5 practical-null verdicts therefore fail (\textsc{uninformative\_incompetent}), and neither directional gate passes (C4: 5/12 margins; C5: 7/12 margins with two reversals). Under the frozen initialization-grouped report, three of the six N$+$ initialization clusters have different phenotypes across their two orders (R$-$/R$+$: 6/6 concordant; G$-$/G$+$: 5/6). Together, these preregistered outcomes leave the C4/C5 decomposition inconclusive. Descriptively, N$+$ shows that high compositional accuracy is attainable with global attention geometry when foreign evidence is replaced by the specified neutral filler; the manipulation does not identify how much of the masking advantage is attributable to foreign task content rather than other consequences of filler replacement.

\paragraph{Communication dependence.} All masked trajectories have full-bank all-cut accuracy exactly at chance ($0.0588$); other competent trajectories fall to near-chance accuracy and satisfy the frozen all-cut bound. Channel necessity in this sense does not establish that the channel is the exclusive source of answer-relevant information, particularly under global evidence access.

\section{Mechanism and dialect audits (non-gating)}\label{sec:mechanism}

After the verdict locked, we ran the preregistered non-gating mechanism battery: thirty audit jobs covering forty-two distinct trajectories --- full cross-regime dialect matrices on all twelve R$+$/G$+$ marked pairs (twenty-four trajectories), plus self-audits on eighteen further trajectories (six R$-$, six G$-$, six N$+$, selected by the frozen lower-order-seed rule, not by performance). The audits intervene at the packet level on initially correct recipient cases from a frozen mechanism bank (24 programs $\times$ 17 values $\times$ 2 diagnostic rotations $=816$ initial episodes; this differs from the behavioral and S1 banks), using the frozen cohort intervention cells: intact replay, same-value substitution under the frozen native donor-sampling rule (exact-episode and program overlap counts are recorded), counterfactual substitution ($+3$ value offset), matched evidence counterfactual (rewriting the start-value span to induce the same intermediate-value change as the packet counterfactual, scored against the matched counterfactual answer), zero-packet replacement (the frozen ``deletion'' cell), random different-value donor substitution sampled with replacement (the frozen ``deranged'' cell), and noise. Donor pools retain at most six initially successful episodes per value and interface in frozen bank order; eligibility also requires the prescribed donor types to be available. Because cases are success-conditioned, the intact rate is 1.000 by construction and is not a necessity statistic; channel dependence is assessed by the deletion cell and the behavioral all-cut. Matrix comparisons use coverage-intersected, success-conditioned case sets, so eligible denominators vary by pair and interface even though the initial bank is identical (per-interface denominators are in the released per-case records). Table~\ref{tab:mechanism} summarizes.

\begin{table}[t]
\centering
\setlength{\belowcaptionskip}{8pt}
\caption{\textbf{Mechanism audit summary by condition.} Self-audit rows: medians over six societies of per-society means across the three relay interfaces. Matrix rows: same-value and cf-follow give ranges of trajectory-wise minima across interfaces; deletion gives the range over all 36 recipient--interface rates per arm. The N$+$ row includes three trajectories with fewer than 30 eligible cases at every interface (7--26); their scores are descriptive and do not support interface classification. See \S\ref{sec:mechanism} for the heterogeneous trajectory results. \emph{Same-value}: Does replacing a packet with a donor packet associated with the same intermediate value preserve the original answer? Donors follow the frozen native sampling procedure. \emph{Cf-follow}: Does replacing a packet with one associated with a different intermediate value produce the predicted changed answer? The donor value is offset by $+3$ modulo 17. \emph{Matched-ev.\ cf}: Does changing the input evidence to induce that same intermediate-value change produce the predicted answer? We rewrite the start-value span and let the system generate its packets normally. \emph{Deletion}: How often is the original answer still correct after replacing the packet with zeros? Low accuracy indicates sensitivity to that intervention on eligible cases, with $1/17$ as a reference chance level. \emph{Shift-retain}: How often does the system keep its original answer under the same packet substitution used for cf-follow? This measures original-answer retention, not a direct assay of whether the packet was internally ignored. $^{\dagger}$Intact is 1.000 by construction (success-conditioned case selection). $^{\ast}$Trajectory-wise minima; individual G$+$ interfaces range higher (one reaches same-value 0.966 with cf-follow 0.017 --- why both metrics matter).}
\label{tab:mechanism}
\footnotesize
\setlength{\tabcolsep}{5.5pt}
\begin{tabular}{lcccccc}
\toprule
Condition & Intact$^{\dagger}$ & Same-value & Cf-follow & Matched-ev.\ cf & Deletion & Shift-retain \\
\midrule
R$-$ (6 self-audits)  & 1.000 & 1.000 & 1.000 & 1.000 & 0.059 & 0.000 \\
R$+$ (12 matrix recipients) & 1.000 & 1.000 & 1.000 & --    & 0.035--0.076 & -- \\
G$-$ (6 self-audits)  & 1.000 & 0.448 & 0.104 & 0.120 & 0.090 & 0.170 \\
G$+$ (12 matrix recipients) & 1.000 & 0.157--0.643$^{\ast}$ & 0.000--0.168$^{\ast}$ & -- & 0.000--0.175 & -- \\
N$+$ (6 self-audits)  & 1.000 & 0.967 & 0.533 & 0.472 & 0.061 & 0.017 \\
\bottomrule
\end{tabular}

\end{table}

\paragraph{All audited masked trajectories meet the value-level intervention criterion.} All eighteen audited masked trajectories --- twelve R$+$ matrix recipients and the six preselected R$-$ self-audits --- achieved same-value preservation and counterfactual following of 1.000 at each tested interface, on the corresponding eligible case sets, with deletion at or near chance and matched-evidence counterfactual following 1.000 (R$-$). This extends the parent study's post hoc-selected, approximately value-indexed audits to every masked trajectory tested here, without exception; it covers eighteen of the twenty-four masked cohort trajectories. Packet substitutions have the expected value-level causal effect within the tested recipient, task, and native-context conditions. This establishes neither a unique or context-free representation nor mediation of the masking advantage.

\paragraph{No audited marked-global trajectory meets strong canonicality at every interface.} No marked global trajectory met strong canonicality (same-value and cf-follow $\geq0.90$) across all three interfaces; eleven of twelve had at least one interface satisfying the preregistered operational context-entanglement indicator (same-value $<0.60$ with deletion $\leq0.70$). The G$-$ self-audits, reported separately with the stated aggregation order, give median same-value 0.448, counterfactual following 0.104, and matched-evidence counterfactual 0.120. These indicators are recipient--interface-level operational descriptions, not exhaustive society-level semantic diagnoses; unlike the masked recipients, the audited marked-global trajectories did not attain strong value-level intervention fidelity across all tested interfaces.

\paragraph{Audited N$+$ trajectories: fidelity tracks competence, at the audit's resolution.} Among the six preselected N$+$ self-audits, two competent trajectories met the strong-canonicality criterion at every interface: one was perfect, and the other scored $815/816$ on both tests at one interface and $816/816$ at the others; a third showed high fidelity but missed the strong-canonicality threshold at one interface (counterfactual following $655/730=0.897<0.90$). The unsuccessful trajectories yielded very small eligible sets (7--26 cases per interface) and are reported descriptively without interface-type classification, per the frozen low-denominator rule ($<30$). The evidence supports an association between successful N$+$ behavior and high value-level intervention fidelity; it does not identify a mechanism of enforcement or mediation, and only six of twelve N$+$ trajectories were audited.

\paragraph{Cross-regime interoperability fails the preregistered conjunction in 11/12 pairs.} Interoperability is distinct from strong canonicality. Classification requires $\geq30$ eligible cases per self/cross test. At each interface, both donor directions must achieve same-value and counterfactual-following scores $\geq0.60$ in self-tests and $\geq0.50$ in cross-regime tests, with both cross/self ratios defined and $\geq0.50$. For each metric and interface, the cross/self ratio divides the cross-regime score by the same recipient's self-substitution score on the matched eligible cases. Eleven of twelve marked pairs had no interface meeting the preregistered bidirectional interoperability conjunction. In the exception (initialization \soc{m1276321262}, order \soc{o3715708347}), only the final interface qualifies: same-value transfer $\approx0.592$/$0.599$ and counterfactual following $\approx0.575$/$0.626$ in the two directions; earlier interfaces do not show the same bidirectional compatibility. Individual directions elsewhere can be substantially better than their reverses (one interface of \soc{m1276321262}\allowbreak\soc{\_o2590280901} reaches R$+{\to}$G$+$ same-value $0.684$ with cf-follow $0.721$ while its reverse is much weaker), so these are failures of the bidirectional conjunction, not proof of unintelligibility in every direction. Poor native-coordinate interoperability between separately optimized matched trajectories does not by itself establish semantically different kinds of code; the companion study documents private dialects even among masked societies \citep{marincat2026companion}.

\section{Discussion}\label{sec:discussion}

\paragraph{What the confirmation establishes.} Under the tested seeds, worlds, and budget: the parent cohort (ten matched pairs, five initializations, one world) showed the masking advantage but formally missed its battery; the present confirmation (twelve matched initialization--order sets, six clusters, a fresh world) passes both the marker-present primary contrast and the unmarked replication in full, with every paired comparison individually exceeding the frozen margin at both depths, and with mechanism audits showing a consistent value-interface difference between regimes. The two cohorts are reported separately; their trajectory counts are not pooled.

\paragraph{Evidence access, text exposure, and content.} The five-arm design distinguishes attention reach (can the cell attend elsewhere?), text exposure (is there foreign text?), and original evidence-token exposure (is that text the original task description?). The N$+$ arm holds reach and span lengths fixed and replaces original foreign evidence tokens with specified neutral filler: seven N$+$ trajectories met the full-generalizer definition, whereas no G$+$ trajectory did; mechanistic comparisons are limited to the preregistered audit sample and its eligible case sets. This is consistent with the hypothesis that access to task-relevant foreign evidence --- the whole-program shortcut --- is the operative ingredient the mask removes; it does not establish that attribution, because filler replacement also changes lexical content and may make the genuine evidence span recognizable, and the preregistered decomposition gates were inconclusive.

\paragraph{Why the markers may have been ignorable.} During training, a cell's own span occupies a fixed slot, so ownership markers are positionally redundant: attention can index spans by position without reading the marker token. This is a plausible account of the marker results --- practical redundancy under masking and failed permutation tracking under visibility --- but the audits do not identify the internal learning mechanism. The documented shuffled-slot follow-up is intended to remove fixed-slot ownership cues and make marker following informative; functional role following would still need to be validated there before any role-rescue conclusion.

\paragraph{The N$+$ split as an optimization observation.} One hypothesis consistent with the census is that masking changes which solutions are readily discovered, while filler replacement provides a less reliable route to similar behavior: seven N$+$ trajectories reached full generalization, five did not (four acquiring little; one only shallowly). We advance this explicitly as a hypothesis; basin structure, shortcut adoption, and learned span-selective attention were not measured.

\paragraph{Implication for modular learning.} The result motivates testing evidence boundaries as a training design choice: a module may learn a more reusable local computation when it cannot directly read the rest of the program. Whether this improves larger models or less structured tasks remains an empirical question. The present result concerns a small affine-composition world and does not establish a general recipe for language-model systems.

\paragraph{Idle-relay pressure.} One possible explanation is that forwarding cells, which have no new operation to apply, encourage the system to preserve intermediate values in its packets. These audits do not directly test whether such idle-relay pressure contributes to learning the observed codes: the mechanism bank's depth-three programs occupy the four cells with a value and three operators, and no experiment varied the proportion of forwarding cells during training. The documented idle-fraction follow-up remains necessary to assess this explanation; we note only that global-regime cells face comparable slot geometry without attaining the same across-interface fidelity profile in the reported audits.

\section{Limitations}\label{sec:limits}

The relay topology, normalized packet anchor, and explicit singleton/forwarding curriculum scaffold the task; this is not a test of composition emerging without that scaffolding. One task world per study (two across the program), one backbone size (0.5B parameters), within-grammar held-out phrasings rather than free language, and a 17-state carrier space. The marker manipulation is thin --- two constant strings --- and positionally redundant during training; the shuffled-slot design is required before any claim about role information proper. With six initialization clusters, the attainable sign-flip floor is $p=0.03125$; confirmatory force rests on the preregistered conjunction gates and effect sizes. The N$+$ filler, while token-length-matched under a committed pseudorandom assignment, is lexically distinguishable from real evidence, so N$+$ societies could in principle learn ``ignore filler'' as a surface cue; length matching also preserves task-correlated span-length cues, so N$+$ is not an information-free control; the decomposition contrasts are inconclusive and the content-based reading is a hypothesis. Mechanism audits are success-conditioned and compare regimes of very different overall competence; they do not establish that canonicality mediates the generalization advantage. Self-audits cover the frozen lower-order-seed half of the R$-$/G$-$/N$+$ arms, not a census. Primary outcomes became available before cohort training was complete, contrary to the planned cohort-wide evaluation embargo; although no outcome-dependent design change is reported, the intended protection against early-outcome influence was not maintained, and missing host-admission transcripts and unretained duplicate audit outputs also limit retrospective verification. The C2 heterogeneity (four large depth-three improvements across three clusters) and the single partial-interoperability interface are descriptive observations this design cannot resolve at $n=6$ clusters.

\section{Documented follow-up proposals}\label{sec:followups}

We propose four follow-ups: (i) \emph{shuffled-slot markers} --- randomize span-to-slot assignment per episode, intended to remove fixed-slot ownership cues and make marker following informative (responds to the preregistration's residual-confound register); (ii) \emph{idle-fraction manipulation} --- vary the number of evidence-bearing cells to test the idle-relay account directly; (iii) \emph{N$+$ fragility} --- what determines which neutral-filler societies reach full generalization; (iv) \emph{iterated learning} --- whether canonical codes sharpen or drift across society generations. The dated build ledger records the idle-fraction and iterated-learning proposals (\texttt{BUILD\_STATUS\_LEDGER\_SNAPSHOT\_2026-09-13.md} in the evidence repository); the shuffled-slot and N$+$ fragility proposals are presented here without a claim of pre-outcome registration.

\section*{Reproducibility and artifact release}

The public confirmation package comprises the frozen preregistration, code inventory and SHA-256 hashes; original and corrected manifests with the denominator-correction record and commit-level chronology; raw evaluation counts for all sixty societies; marker-permutation and mechanism/dialect audit records; secondary-statistics outputs; the machine-admission receipt index; the one-shot verdict lock and process archive; and the incident ledger. These original deposits are available in the evidence repository, \url{https://github.com/tokenosopher/populus-evidence-partitioning}, under \texttt{cohort\_\allowbreak prereg/}; audit and receipt records are pinned at revision \href{https://github.com/tokenosopher/populus-evidence-partitioning/tree/ebe2ebf80d55d52a6ef9b6c237e2fee1e6c99294/cohort_prereg}{\texttt{ebe2ebf}}. All sixty final step-20{,}000 checkpoints and a SHA-256 inventory are deposited at \url{https://huggingface.co/tokenosopher/populus-evidence-partitioning-checkpoints}, under \texttt{cohort\_\allowbreak prereg/}, revision \texttt{a3b066e7}. The manuscript source archive rebuilds this paper; it does not include model checkpoints or claim an end-to-end training reproduction.

Historical manuscript-review and validation records for v5 are retained in the \href{https://github.com/tokenosopher/populus-evidence-partitioning/tree/3433fcdc6d53cba10bdb621ea46ccea6e2c464f2/cohort_prereg/manuscript_review/v5}{public v5 review deposit}, preserved byte-for-byte from local paper commit \texttt{200d415}. Those records concern earlier drafts and AI-assisted preparation, not independent peer review of this paper. The public release also includes the duplicate-execution disclosure and the linked historical audit-summary erratum (\texttt{ERRATUM\_AUDIT\_PHASE\_SUMMARY.md}). Duplicate audit execution used the disclosed first-harvested selection rule; later self-audit copies were not retained, so agreement between competing copies could not be assessed. Deterministic settings are not a substitute for that unperformed comparison. Training and evaluation used enforced deterministic-algorithm settings with TF32 disabled; all reported behavioral evaluations and mechanism audits use final step-20{,}000 checkpoints with no checkpoint selection.

\section*{Funding and competing interests}

This research was personally funded by the author. No external research funding was received. The author declares no relevant competing interests.

\section*{Use of generative AI}

The author originated the research ideas and experimental design and developed and oversaw the end-to-end experimental workflow, including monitoring experimental outputs. Generative AI tools were used for low-level coding and implementation of those ideas, building and operating the training and evaluation pipeline, monitoring, result reporting and consistency checks, and drafting, reviewing, and revising the manuscript. The author personally reviewed and verified the reported results using this workflow and AI-assisted checks. AI also contributed to protocol refinement through the adversarial review described in Appendix~\ref{app:integrity}; its written rulings were incorporated as procedural constraints. These uses do not constitute independent peer review or independent replication. The author takes responsibility for the experimental design, implementation, analysis, manuscript, and claims. The verification and record-retention limitations described in \S\ref{sec:integrity} and Appendices~\ref{app:integrity}--\ref{app:correction} remain applicable.

\bibliographystyle{tmlr}
\bibliography{references}

\clearpage
\appendix

\section{Protocol and integrity records}\label{app:integrity}

\paragraph{Freeze before seed, seed before training.} All code, gate thresholds, contrast definitions, phenotype classifiers, neutral grammar, filler banks, and the verdict script were frozen and hash-inventoried (a SHA-256 recorded for each of the 38 files, making any later modification detectable) before any seed generation; the freeze deposit was pushed to the public evidence repository and independently archived by Software Heritage (snapshot \texttt{swh:1:snp:\allowbreak b3fd6afb\allowbreak 0b1825d5\allowbreak 95189449\allowbreak 35e8c8b3\allowbreak ddbf38a8}, 2026-09-11) before the master seed was drawn. World seeds derive deterministically from a committed master string; a candidate world is accepted only by frozen structural checks, and every operator seed previously exposed during development is excluded by a committed denylist. Frozen constructors and rules are distinguished from world-dependent artifacts, which were instantiated only after seed generation. The populated pre-run manifest --- the pre-run paperwork recording expected file hashes, evaluation-bank sizes, and seeds --- was likewise deposited and archived (snapshot \texttt{swh:1:snp:\allowbreak 169891a9\allowbreak e52e540e\allowbreak ab85eea2\allowbreak a6649f4e\allowbreak b1923ab6}) before training was permitted.

\paragraph{Adversarial referee.} The author used a separate AI conversation tasked with adversarial protocol review (GPT-5.6 Pro, per the retained session record); its written rulings were incorporated as binding procedural constraints, and training was authorized only on an explicit GO ruling. This was AI-assisted methodological review, not external institutional replication or conventional journal peer review; responsibility for the protocol and claims remains with the author. All five original rulings are published verbatim in the evidence repository.

\paragraph{Machine admission.} Every rented GPU instance had to pass four gates before contributing cohort training (the golden retrace itself performs training updates): (i) a hardware health battery (GPU, driver, and CUDA functional checks); (ii) an environment install in which every downloaded artifact was verified by size and checksum; (iii) a \emph{golden retrace} --- the machine re-runs a short reference training segment and must reproduce the stored reference outputs exactly, bit for bit, demonstrating agreement on that tested prefix, not equivalence for every computation; and (iv) a mask-verification test run on that machine. The mask test has two directions. The leak test: rewriting the tokens of a span the cell is masked away from must produce exactly zero change --- bit for bit --- at the tested writer residual where the cell writes its outgoing packet (incoming packets held fixed, so only the mask is being tested); information that cannot reach a cell cannot change its output, so any difference means the mask leaks and the machine is rejected. Complementary checks test source-cell own-span liveness, source-cell foreign-span liveness under global visibility, and downstream incoming-packet reader liveness; attention-row assertions check the specified question, span, and packet-readiness layout (nine Boolean acceptance fields combining perturbation checks, attention-layout checks, and a code-level assertion). These are finite implementation tests at the stated perturbations and locations. During training, each society emitted running checksums of its training-example stream, plus a \emph{state fingerprint} --- one hash covering model weights, optimizer state, random-number-generator states, and the step counter --- taken after its first three updates. For one society, this three-update fingerprint was additionally reproduced bit-exactly on a second host admitted through the same pipeline; that cross-host comparison covers only the first three updates and was not an end-to-end reproduction. A receipt index accompanies the release: the per-society three-update fingerprints are recovered from retained logs for all sixty societies, while the transcripts of the health, retrace, and mask tests on the fleet hosts were not retained --- those checks gated every launch, but their execution cannot be independently re-verified from the released record. This is disclosed as a provenance limitation, not offered as independently verified admission.

\paragraph{One-shot verdict.} The verdict script ran exactly once, under an exclusive lock file (created atomically, so another invocation using the same intact lock is refused), on the compiled results bundle (SHA-256 \texttt{7a8a920b\ldots}) and the corrected manifest (SHA-256 \texttt{7632aa76\ldots}); the complete process record (command, interpreter, platform, input hashes, stdout) is archived. The post-evaluation, pre-verdict denominator correction and the deviation from the cohort-wide evaluation embargo are documented in Appendix~\ref{app:correction}, including the original failed compiler invocation, bounded correction procedure, outcome exposure, and commit-level chronology.

\section{The manifest correction, in full}\label{app:correction}

During result compilation, the compiler's first integrity gate fail-closed (aborted rather than proceed) on a mismatch between the pre-run manifest's expected single-operation-diagnostic denominator (1{,}224) and the frozen evaluator's output (2{,}448). Diagnosis: the manifest-population expression used $12\times3\times17\times2$ --- retaining a stale two-rotation factor --- where the frozen evaluation specification fixes four rotations ($12\times3\times17\times4=2{,}448$). A dataflow check (deposited) established that the stale value was consumed only by the compiler's consistency gate and the manifest validator, never by training, evaluation-case selection, or score normalization; all sixty stored S1 scores were computed over the frozen four-rotation bank. The AI protocol referee ratified a bounded two-field correction (\texttt{expected\_denominators.S1} and \texttt{bank\_sizes.S1}, $1{,}224\to2{,}448$) classified as a \emph{post-evaluation, pre-verdict administrative correction to derived manifest metadata}, with binding provenance requirements (original manifest preserved byte-for-byte under its immutable commit; machine-checked structural diff confirming no other field changed; full chronology including the aborted compiler invocation; no claim that the aborted invocation computed any verdict quantity). That ruling was explicitly conditional on the submitted facts and recorded its own verification boundary; it is not independent certification of subsequent execution.

\paragraph{Chronology (commit-level).} The correction record's narrative timestamps (``$\sim$20:0x BST'') were approximate; the recorded Git and process timestamps (supplied metadata preserved in the immutable history, not third-party upload receipts) give the order: correction deposit commit \texttt{b8977cf}, 2026-09-12 19:58:03 BST; verdict process start 18:58:36 UTC $=$ 19:58:36 BST (33 seconds after the correction deposit); verdict and bundle deposit commit \texttt{8358ed0}, 20:00:28 BST. A chronology addendum reconciling the narrative note against these receipts accompanies the correction record; the original note is preserved unmodified.

\paragraph{Outcome exposure and evaluation timing.} The preregistration permitted an isolated liveness bank during training and required the primary final bank to be first evaluated only after all designated step-20{,}000 checkpoints completed. The retained file-time receipts show a deviation: the lane pipeline executed each society's frozen primary evaluation immediately after that society's training completed, so 59 of 60 primary evaluation outputs predate the last checkpoint's completion. These were final-bank evaluation executions, not liveness monitoring, and are not relabelled as such. Scope: the evaluations are deterministic, frozen, and final-checkpoint-only; the tier was committed at freeze; no outcome-dependent stopping, arm selection, audit selection, or scientific protocol amendment is reported, and the committed roster, final-checkpoint rule, and frozen analysis definitions were retained; reruns occurred only on objectively logged infrastructure failures. Descriptive per-society summaries (depth accuracies, all-cut, marker-audit rates) were consequently inspected before the correction and verdict: the correction was pre-verdict but not outcome-blind. Its mechanical justification (the frozen four-rotation specification and the hash-matching frozen evaluator, which determine 2{,}448 independently of any performance result) does not rest on a blinding claim.

\paragraph{Execution sequence.} One result-compiler invocation aborted at its first integrity gate before computing any verdict, contrast decision, or phenotype classification; following deposit of the ratified correction, a subsequent invocation of the unchanged compiler compiled the unchanged sixty-society records against the corrected manifest; the separate verdict program was then executed once under the exclusive lock.

\end{document}